\documentclass{article}

\usepackage[preprint]{neurips_2025}

\usepackage[utf8]{inputenc}
\usepackage[T1]{fontenc}
\usepackage{newtxtext,newtxmath}
\usepackage[hypertexnames=false]{hyperref}
\usepackage{url}
\usepackage{booktabs}
\usepackage{array}
\usepackage{amsmath}
\usepackage{microtype}
\usepackage[table]{xcolor}
\usepackage{graphicx}
\usepackage{multirow}
\usepackage{algorithm}
\usepackage{algorithmic}
\usepackage{placeins}
\usepackage{cleveref}
\usepackage{caption}
\usepackage{subcaption}



\title{OneLatent: Single-Token Compression for Visual Latent Reasoning}

\author{
  Bo Lv$^{1*\P}$, Yasheng Sun$^{2*}$, Junjie Wang$^{1\dagger}$, Haoxiang Shi$^{3\dagger}$ \\
  $^{1}$Tsinghua University \\
  $^{2}$Institute of Science Tokyo \\
  $^{3}$Waseda University \\
  \small{\texttt{boironic@gmail.com, sunyasheng01@yeah.net,}} \\
  \small{\texttt{wangjunjie@sz.tsinghua.edu.cn, hollis.shi@toki.waseda.jp}}
}

\newcommand{\tokbot}{\texttt{<|begin-latent|>}}
\newcommand{\tokeot}{\texttt{<|end-latent|>}}
\newcommand{\toklat}{\texttt{<|latent|>}}

\begin{document}

\maketitle

\begin{abstract}
Chain-of-thought (CoT) prompting improves reasoning but often increases inference cost by one to two orders of magnitude. To address these challenges, we present \textbf{OneLatent}, a framework that compresses intermediate reasoning into a single latent token via supervision from rendered CoT images and DeepSeek-OCR hidden states. By rendering textual steps into images, we obtain a deterministic supervision signal that can be inspected and audited without requiring the model to output verbose textual rationales. Across benchmarks, OneLatent reduces average output length by $11\times$ with only a $2.21\%$ average accuracy drop relative to textual CoT, while improving output token contribution (OTC) by $6.8\times$. On long-chain logical reasoning, OneLatent reaches $99.80\%$ on ProntoQA and $97.80\%$ on ProsQA with one latent token, with compression up to $87.4\times$, supporting compression-constrained generalization.
\end{abstract}
\renewcommand{\thefootnote}{}
\footnotetext[1]{*\hspace{0.2em}Equal contribution}
\footnotetext[2]{\dag\hspace{0.2em}Corresponding author}
\footnotetext[3]{\P\hspace{0.2em}This work was performed when Bo Lv was visiting Tsinghua University as a research intern.}
\renewcommand{\thefootnote}{\arabic{footnote}}

\section{Introduction}
Explicit chain-of-thought (CoT) prompting is one of the most effective ways to improve reasoning in language models ~\citep{wei2022chain,wang2022self,zhou2022least,khot2022decomposed,wang2022towards,sprague2024cot}, but it is expensive at deployment: output length, latency, and KV-cache usage all grow with generated reasoning tokens. More importantly, long traces are not always high-information traces; they often mix useful computation with template-like verbal redundancy ~\citep{turpin2024language}.

This motivates an minimum description length (MDL) view ~\citep{Gr_nwald_2019}: if two intermediate explanations both produce the correct answer, the shorter sufficient one can provide a stronger inductive bias and reduce redundant verbal scaffolding. However, MDL is not directly optimized in current CoT pipelines, which are trained with token-level likelihood on explicit traces rather than description length of internal reasoning states.

The challenge is that naive compression fails: weak compression keeps redundant verbal scaffolds, while aggressive compression removes task-critical computation. We therefore introduce a visual latent token: a learnable continuous latent slot whose representation is aligned to DeepSeek-OCR vision-encoder hidden-state targets derived from rendered CoT images ~\citep{wei2025deepseekocrcontextsopticalcompression}. This differs from prior latent-reasoning approaches that stay in text space and typically allocate multiple latent tokens to carry intermediate computation. Importantly, the visual latent token is not an image patch token at inference; test-time inputs contain only the question and a single latent slot before answer generation.

\begin{figure}[t]
  \centering
  \includegraphics[width=\linewidth]{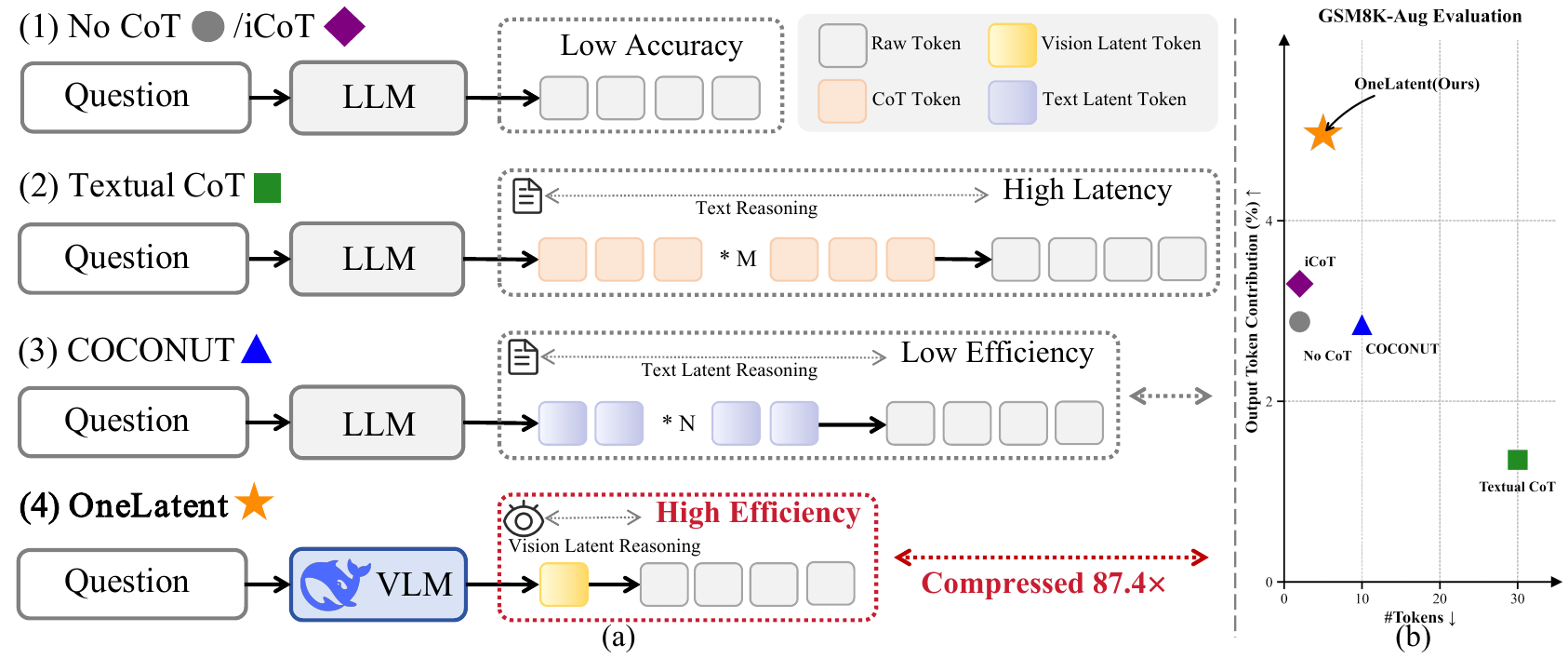}
  \caption{Reasoning interfaces and efficiency plot. (a) We illustrate four reasoning interfaces and their output forms: No CoT, textual CoT, iCoT, COCONUT, and OneLatent. (b) On GSM8K, we plot OTC (\%/token) versus the average number of generated output tokens for the same methods. OneLatent appears in the lowest-output-length region and keeps high OTC, indicating substantially shorter outputs than the compared reasoning interfaces.}
  \label{fig:comparison}
\end{figure}

This aligns with the MDL intuition: under a strict intermediate-interface budget, the model is encouraged to represent only what is sufficient for answering. We do not optimize MDL explicitly; instead, we impose a single-latent interface and evaluate the resulting accuracy--length trade-off. Figure~\ref{fig:comparison} (b) shows the resulting trade-off versus No-CoT, textual CoT, iCoT, COCONUT and OneLatent.

To quantify efficiency under constrained decoding, we introduce output token contribution (OTC), defined as $\mathrm{OTC}=\mathrm{Acc}/\mathrm{AvgOut}$, where $\mathrm{Acc}$ is accuracy in percent and $\mathrm{AvgOut}$ is average generated output tokens (\%/\text{token}). Unless noted otherwise, accuracy is macro-averaged across five benchmarks (GSM8K, GSM8K-Hard, SVAMP, ProntoQA, and ProsQA), and output length counts generated tokens excluding the prompt under identical decoding settings. Under this protocol, OneLatent reduces average output length from $74.62$ to $6.78$ tokens and incurs only a $2.21\%$ macro-averaged accuracy drop relative to textual CoT, while improving OTC by $6.8\times$; in long-CoT settings, compression reaches up to $87.4\times$, with degradation concentrated in arithmetic-heavy tasks.

The main contributions are threefold:
\begin{itemize}
  \item We propose a \textbf{visual-latent compression method} that uses DeepSeek-OCR hidden states to supervise one latent token with a stable three-stage curriculum.
  \item We propose an MDL-motivated \textbf{data preparation and CoT rendering strategy} that deterministically renders variable-length CoT into fixed-size images and compresses them into stable visual hidden-state targets for single-token latent supervision.
  \item We provide \textbf{evidence for compression-constrained generalization}: $11\times$ output reduction with small accuracy loss, strong long-chain performance, and large gains in OTC as an MDL-motivated efficiency proxy.
\end{itemize}

\section{Related Work}

\paragraph{Latent reasoning methods.}
% CODI citation temporarily removed per current draft scope:
% shen2025codicompressingchainofthoughtcontinuous
Recent work internalizes reasoning into hidden states to reduce token-level deliberation. iCoT ~\citep{deng2023implicit,deng2024explicit}, COCONUT ~\citep{hao2025traininglargelanguagemodels}, and related pause/latent-compute methods ~\citep{goyal2023think,zelikman2024quiet,pfau2024let,wei2025simcotsupervisedimplicitchainofthought} demonstrate that latent computation is feasible. However, these methods often show an accuracy gap relative to explicit CoT and can be sensitive to multi-stage optimization ~\citep{hao2025traininglargelanguagemodels}. Our approach keeps the latent interface minimal (single token) and uses external visual supervision to provide an inspectable, high-density target for guiding single-token distillation.

\paragraph{Text-based reasoning methods.}
Textual CoT and its variants remain strong baselines for multi-step reasoning ~\citep{wei2022chain,wang2022self,zhou2022least,khot2022decomposed,wang2022towards,madaan2022text,sprague2024cot}. Follow-up work improves reliability via verification, self-refinement, and search/planning at inference time ~\citep{shinn2023reflexion,madaan2023self,wang2024math,xie2023self,hao2023reasoning,yao2023tree,hao2024llm,wang2023guiding,valmeekam2023planning,welleck2024decoding}. A key limitation is decoding cost: stronger performance often requires longer textual traces, and these traces are not always reliable indicators of the model's internal computation and may include post-hoc or redundant content ~\citep{turpin2024language}. OneLatent uses textual CoT as \emph{training supervision} but removes it from the runtime interface.

\paragraph{Visual context compression.}
Visual compression methods encode long text segments as compact image representations. Glyph ~\citep{cheng2025glyphscalingcontextwindows} and DeepSeek-OCR ~\citep{wei2025deepseekocrcontextsopticalcompression}, together with strong vision-language encoders ~\citep{radford2021learningtransferablevisualmodels}, show that visual channels can preserve dense textual semantics under aggressive compression. Prior work mainly targets context scaling or OCR quality, not latent reasoning supervision. We instead use optical compression as a bridge from long CoT traces to a single reasoning token, enabling compact inference while retaining much of the reasoning performance.

\section{OneLatent Method}
\label{sec:method}
\subsection{Overview}
We transform explicit chain-of-thought reasoning into a single compressed latent token through a three-stage training curriculum: (1) explicit CoT Cold Start, (2) visual latent supervision via rendered CoT images, and (3) answer-only fine-tuning. We call this slot the visual latent token: the model learns to align this continuous slot with hidden-state targets extracted by a frozen vision-encoder from rendered CoT images.

\begin{figure}[t]
  \centering
  \includegraphics[width=\linewidth]{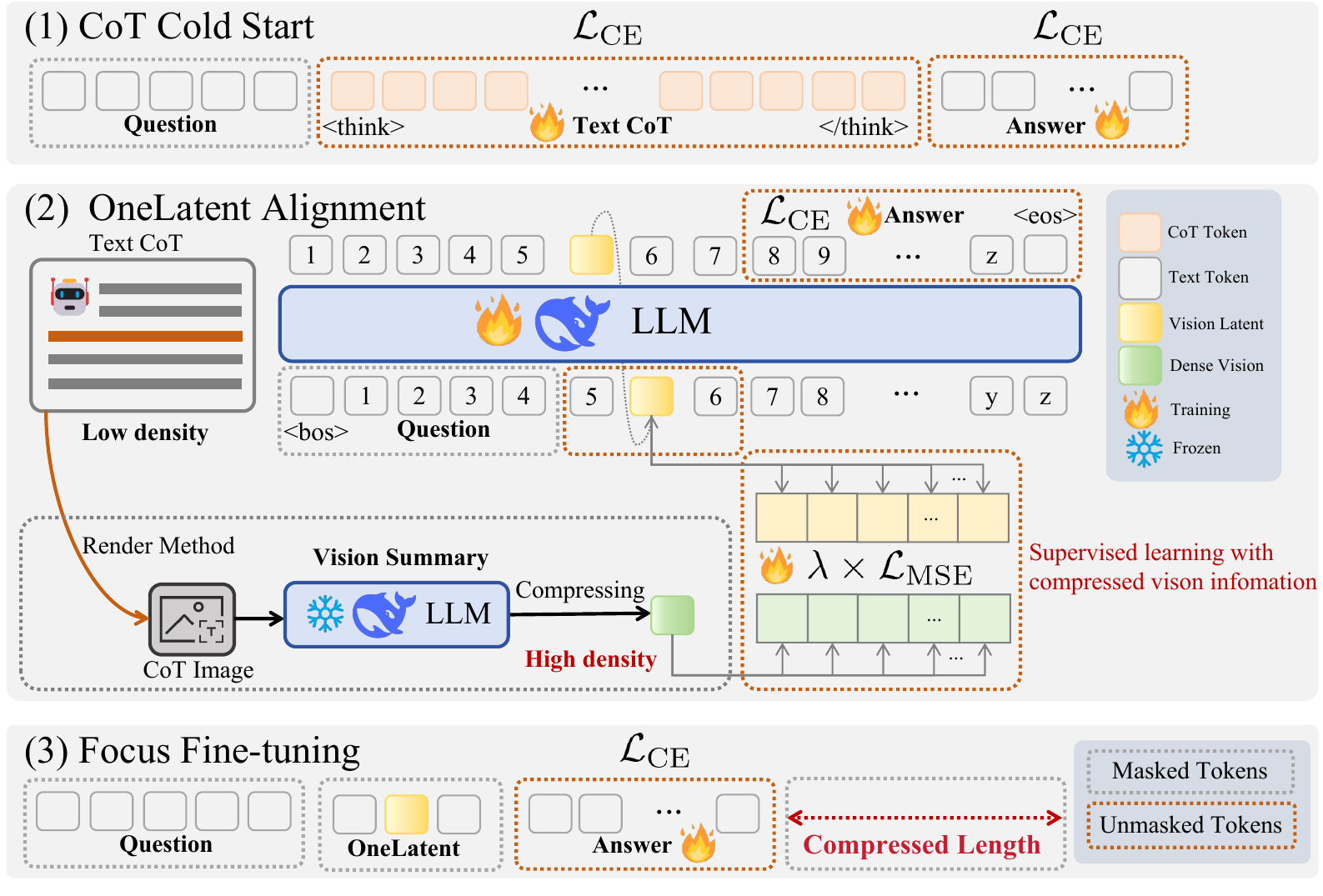}
  \caption{Three-stage training pipeline of OneLatent. Stage 1 trains explicit CoT generation with cross-entropy. Stage 2 replaces CoT with one latent token and adds MSE alignment to pre-extracted visual hidden-state targets. Stage 3 keeps the one-latent interface and trains answer generation without alignment loss. The figure shows the supervision shift from explicit textual CoT to latent alignment and then to answer-only decoding.}
  \label{fig:main-pipeline}
\end{figure}

Figure~\ref{fig:main-pipeline} illustrates the complete OneLatent pipeline, which consists of two main phases: an offline data preparation phase and an online three-stage training phase. Our target extraction follows the DeepSeek-OCR architecture, which includes a vision-encoder and an LLM backbone. In the data preparation phase, we render explicit CoT text into images using a deterministic layout algorithm, then extract hidden-state targets by passing these images through this frozen DeepSeek-OCR stack. These precomputed targets serve as supervision signals during training. In the three-stage training phase, we progressively compress reasoning: Stage 1 teaches the model to generate explicit CoT reasoning, Stage 2 aligns a single latent token to the visual hidden-state targets while removing explicit CoT generation, and Stage 3 fine-tunes answer generation to consolidate latent reasoning without alignment loss. At inference time, the model uses only the latent token to compress reasoning internally, generating short answers without explicit intermediate steps.

\subsection{Latent Segment and Continuous Filling}
Given a question $q$ and answer $a$, we insert a latent segment between them:
\begin{equation}
  x = [\text{BOS}, q, \tokbot, \underbrace{\toklat, \ldots, \toklat}_{N}, \tokeot, a, \text{EOS}].
\end{equation}
For all reported evaluations, we use $N=1$. At latent positions, we overwrite the embedding with the previous hidden state. For latent position $\ell$:
\begin{equation}
  e_{\ell} \leftarrow h_{\ell-1}, \quad h_{\ell} = f_\theta(e_{\le \ell}).
\end{equation}
This creates a continuous hidden-state ``thinking'' process, and the same mechanism is used during training and inference.

\subsection{Data Preparation and CoT Rendering}
Our rendering pipeline follows the \texttt{onelatent\_gsm8k} implementation. Each CoT trace is rendered into a fixed square canvas ($W{=}H{=}1024$ for standard GSM8K-Aug, $W{=}H{=}512$ for tiny runs), with padding $p$ and a mono-spaced font (DejaVu/Liberation/FreeMono fallback). We approximate the maximum characters per line by
\begin{equation}
  m(f) = \left\lfloor \frac{W - 2p}{f/2} \right\rfloor ,
\end{equation}
wrap the text into $L=\lceil |s|/m(f)\rceil$ lines, and render in black on white. To guarantee that the entire CoT fits inside the fixed canvas, we perform a bounded search over font sizes $f \in [f_{\min}, f_{\max}]$ and select the largest size that satisfies the height constraint
\begin{equation}
  L \cdot (f + g) + 2p \le H,
\end{equation}
where $g$ is a line-gap factor (we use $g{=}f/4$). This ``fit-to-canvas'' search is lightweight and deterministic. The result is a single image that preserves the left-to-right reading order while ensuring no truncation.

\begin{figure}[t]
  \centering
  \includegraphics[width=\linewidth]{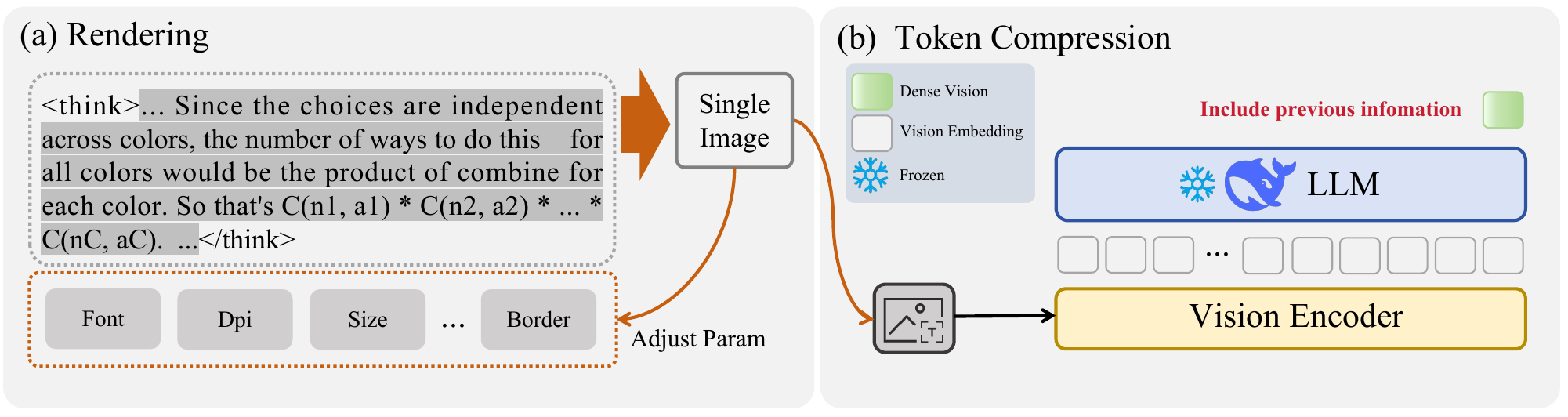}
  \caption{Data preparation pipeline for latent supervision. (a) CoT text is rendered into a fixed-size image with controlled layout parameters. Middle: the rendered image is encoded by frozen visual modules and the LLM backbone. (b) a hidden-state target is extracted and stored for Stage $2$ supervision. The figure depicts one offline target vector produced from each CoT sample.}
  \label{fig:data-prep}
\end{figure}

Figure~\ref{fig:data-prep} illustrates the complete data preparation pipeline. The process begins with the original dataset containing questions, explicit CoT reasoning, and answers. We first render each CoT text into a fixed-size image using our deterministic layout algorithm, which automatically adjusts font size to ensure the entire reasoning trace fits within the canvas while maintaining readability. These rendered images are then processed through the DeepSeek-OCR vision-encoder, which consists of frozen SAM-ViT-B and CLIP-L encoders. The visual features are concatenated, projected to the LLM hidden dimension, and forwarded through all LLM layers. We extract the final hidden state at the last position as the target supervision vector $v \in \mathbb{R}^d$, which is stored as a target hidden state for use during Stage~2 training. This offline preparation ensures that training requires only text input while maintaining a visual anchor for latent alignment.

\paragraph{Quality verification via DeepSeek-OCR.} To ensure rendered images are clear and readable, we employ DeepSeek-OCR~\citep{wei2025deepseekocrcontextsopticalcompression} as an automatic quality checker. For each rendered image, we use the vision-encoder to verify whether the text is legible. If the OCR confidence score falls below a threshold or text recognition accuracy is poor, we automatically adjust rendering parameters including dots per inch(DPI) and border padding. This iterative refinement process continues until the rendered image achieves satisfactory clarity, ensuring that the visual targets extracted from these images provide stable supervision signals for latent alignment.

\paragraph{Rendering algorithm.} Algorithm~\ref{alg:render} summarizes the procedure used in our preprocessing procedure. We first normalize the CoT string by collapsing whitespace, converting LaTeX-like tokens (e.g., \texttt{\textbackslash times}, \texttt{\textbackslash leq}) into Unicode, and stripping markup that is not part of the reasoning trace. We then select a mono-spaced font (with deterministic fallback order) and compute a conservative characters-per-line budget to preserve left-to-right reading order. A descending search over font sizes ensures the full trace fits in the fixed canvas; if no size satisfies the height constraint, we fall back to the minimum font size and clip only trailing whitespace. This deterministic rendering pipeline makes the image targets reproducible across machines and allows caching during large-scale preprocessing.
\begin{algorithm}[t]
\caption{Deterministic CoT Rendering with OCR-Based Quality Check}
\label{alg:render}
\begin{algorithmic}[1]
\STATE \textbf{Input:} CoT string $s$, canvas size $(W,H)$, padding $p$, font range $[f_{\min}, f_{\max}]$, DPI range, quality threshold $\tau$
\STATE \textbf{Initialize:} $\text{DPI} \leftarrow \text{DPI}_{\text{default}}$, $p \leftarrow p_{\text{default}}$
\REPEAT
  \FOR{$f$ from $f_{\max}$ down to $f_{\min}$}
    \STATE $m \leftarrow \lfloor (W-2p)/(f/2) \rfloor$
    \STATE wrap $s$ into $L$ lines using width $m$
    \STATE \textit{Wrap uses greedy whitespace breaks; overlong tokens fall back to character-level splitting.}
    \STATE $h \leftarrow L\cdot(f+g) + 2p$
    \IF{$h \le H$}
      \STATE render image $I$ with font size $f$, DPI, padding $p$
      \STATE \textbf{break}
    \ENDIF
  \ENDFOR
  \STATE $Q \leftarrow \text{DeepSeek-OCR}(I)$ \COMMENT{Compute OCR quality score}
  \IF{$Q < \tau$}
    \STATE Adjust DPI or padding $p$ \COMMENT{Increase clarity}
  \ENDIF
\UNTIL{$Q \ge \tau$ or max iterations reached}
\STATE save image $I$ (white background, black text)
\RETURN $I$
\end{algorithmic}
\end{algorithm}

\subsection{Hidden-State Targets via DeepSeek-OCR Encoder}
We encode rendered CoT images with the DeepSeek-OCR vision-encoder stack ~\citep{wei2025deepseekocrcontextsopticalcompression}. Each image is padded to the target size, passed through the frozen vision-encoder (SAM-ViT-B and CLIP-L) ~\citep{kirillov2023segment,radford2021learningtransferablevisualmodels}, concatenated, and projected to the LLM hidden dimension. We then concatenate a BOS embedding with the visual embeddings and forward through the LLM layers; the final hidden state at the last visual position is used as the target latent vector $v \in \mathbb{R}^d$:
\begin{equation}
  v = \text{LM}\left([\text{BOS}; \text{Vision}(I)]\right)_{-1}.
\end{equation}
For the single-latent regime, each example yields one target vector. The target is stored as \texttt{.pt} and used only for MSE supervision in Stage~2.
\paragraph{Text-only training and inference.} During training and inference, the model never consumes images directly. Instead, precomputed hidden-state targets derived from rendered CoT images are loaded from disk and used only for MSE supervision in Stage~2. This keeps the runtime interface purely textual while retaining a visual anchor during training.
\paragraph{Patch alignment and OCR-friendly layout.} We favor mono-spaced fonts and fixed margins to keep glyph spacing predictable across samples. This allows the vision-encoder to see stable, left-to-right visual patterns that resemble OCR inputs. Since the downstream supervision target is extracted after the LLM layers, minor pixel-level variations are smoothed by the encoder and projector, but large layout shifts can destabilize training. Empirically, fixed-size canvases with deterministic wrapping produce the most stable alignment loss.

\subsection{Three-Stage Training Strategy}
We train with a three-stage progressive strategy that gradually transforms explicit CoT reasoning into compressed latent representations while ensuring answer quality. Stages~1 and~3 follow standard supervised fine-tuning (SFT) on text sequences (next-token prediction), while Stage~2 augments SFT with an additional latent-alignment loss.
\paragraph{Stage 1 (CoT Cold Start).} Input sequence:
\begin{equation*}
  x^{(0)} = [q, \tokbot, \tokeot, r, a],
\end{equation*}
where $r$ is the explicit CoT text. This is an SFT stage: we optimize NTP on CoT + answer tokens and do not use MSE.
\paragraph{Stage 2 (OneLatent Alignment).} Input sequence:
\begin{equation*}
  x^{(1)} = [q, \tokbot, \toklat, \tokeot, a].
\end{equation*}
We optimize NTP on answer tokens and add MSE between the \tokbot hidden state and target $v$ (SFT + alignment).
\paragraph{Stage 3 (Focus Fine-tuning).} Input sequence:
\begin{equation*}
  x^{(2)} = [q, \tokbot, \toklat, \tokeot, a].
\end{equation*}
We drop MSE and continue SFT on answer tokens to consolidate decoding quality.

We optimize next-token prediction (NTP) on the answer tokens and (optionally) CoT tokens:
\begin{equation}
  \mathcal{L}_{\text{NTP}} = - \sum_{t \in \mathcal{A}} \log p_\theta(x_t \mid x_{<t}),
\end{equation}
where $\mathcal{A}$ denotes supervised token positions (CoT+answer in Stage~1; answer-only in Stages~2 and~3). For latent alignment we use:
\begin{equation}
  \mathcal{L}_{\text{MSE}} = \| h_{\text{bot}} - v \|_2^2,
\end{equation}
which compares the hidden state at the \tokbot position to the pre-extracted target. The total loss is:
\begin{equation}
  \mathcal{L} =
  \begin{cases}
    \mathcal{L}_{\text{NTP}}, & \text{Stage 1} \\\\
    \mathcal{L}_{\text{NTP}} + \lambda \mathcal{L}_{\text{MSE}}, & \text{Stage 2} \\\\
    \mathcal{L}_{\text{NTP}}, & \text{Stage 3}
  \end{cases}
\end{equation}
with $\lambda=1$ in our default configuration.

For $N=1$, the MSE aligns the hidden state at the \tokbot position with $v$ and the latent token is filled with this hidden state during forward pass.

\subsection{Inference}
At inference, we construct the same latent segment and generate only the answer. The latent token is filled by the previous hidden state, and decoding proceeds autoregressively from \tokeot.

\paragraph{Fixed latent length design.} We keep a fixed latent length ($N=1$) to maintain a deterministic interface and a one-to-one correspondence between each rendered CoT image and its target hidden state. Dynamic latent length would require a separate length predictor and would decouple the latent supervision target from the latent slot being trained, complicating alignment and degrading training stability. Fixed length also guarantees consistent compute and simplifies batching.
\paragraph{Complexity and efficiency.} Let $T$ be the answer length and $N$ the latent length. Autoregressive decoding cost is $O(T{+}N)$, while explicit CoT incurs $O(T{+}R)$ with $R \gg N$. This is the primary source of inference speedups. Training costs include rendering and visual encoding, but those are offline and do not affect runtime inference.

\FloatBarrier
\section{Experiments}
\subsection{Experimental Setup}
\subsubsection{Datasets}
We separate training sources from evaluation benchmarks. For training, we use GSM8K-Aug-NL~\citep{deng2023implicit}, an augmented GSM8K~\citep{cobbe2021training} corpus with approximately $385$k training samples (used for supervision and latent-target preparation, not as a standalone row in Table~\ref{tab:method-comparison}). For evaluation, we report two settings: (1) five standard reasoning benchmarks (GSM8K, GSM8K-Hard, SVAMP, ProntoQA, ProsQA) in Table~\ref{tab:method-comparison}; and (2) two enhanced long-chain benchmarks (ProntoQA Enhanced, ProsQA Enhanced) in Table~\ref{tab:main-results}.

\subsubsection{Enhanced Dataset Generation}
For enhanced-set evaluation, we use ProntoQA Enhanced (290 test samples, logical reasoning) and ProsQA Enhanced (500 test samples, procedural reasoning). Both contain explicit CoT traces that are rendered into images for visual latent supervision.
\paragraph{ProntoQA/ProsQA enhanced.} We generate enhanced variants by expanding CoT steps using large language models and validating them through LLM-as-Judge verification to ensure both correctness and sufficient reasoning depth. Algorithm~\ref{alg:cot-expansion} describes the expansion and validation procedure.

\begin{algorithm}[t]
\caption{LLM-Based CoT Expansion with Judge Validation}
\label{alg:cot-expansion}
\begin{algorithmic}[1]
\STATE \textbf{Input:} Original CoT text $C_0$, target length $L_{\text{target}}$, max iterations $K$
\STATE \textbf{Output:} Validated expanded CoT $C^*$
\STATE $C \leftarrow C_0$
\FOR{$k = 1$ to $K$}
  \IF{$|C| \ge L_{\text{target}}$}
    \STATE \textbf{break}
  \ENDIF
  \STATE $C' \leftarrow \text{LLM}_{\text{expand}}(C)$ \COMMENT{Prompt LLM to add intermediate steps}
  \STATE $\text{valid} \leftarrow \text{LLM}_{\text{judge}}(C', \text{original\_answer})$ \COMMENT{Verify correctness}
  \IF{$\text{valid}$}
    \STATE $C \leftarrow C'$
  \ELSE
    \STATE \textbf{continue} \COMMENT{Reject invalid expansion, retry}
  \ENDIF
\ENDFOR
\STATE $C^* \leftarrow C$
\RETURN $C^*$
\end{algorithmic}
\end{algorithm}

Our preprocessing procedure converts LaTeX-like tokens to Unicode and renders each expanded CoT into $1024\times1024$ PNG images. Hidden-state targets are then extracted by processing these images through the vision-encoder to produce a single $[1,1280]$ target vector per sample.

\paragraph{GSM8K-Aug for training.} We do not create a separate ``GSM8K-Aug Enhanced'' evaluation table in this paper. GSM8K-Aug-NL is used as a training source (and in-stage ablations), while the enhanced evaluation in Table~\ref{tab:main-results} focuses on ProntoQA/ProsQA where reasoning chains are longer and structurally constrained.

\paragraph{Rendering configuration.} We use $1024\times1024$ PNG images with white background, black text, and mono-spaced font (Courier New). The font size is automatically adjusted via the fit-to-canvas search in Algorithm~\ref{alg:render} to ensure CoT text fits within the image boundaries. Border padding is set to $20$ pixels, and DPI is fixed at $100$ for consistent rendering across all datasets.

\subsubsection{Training Details}
We use the DeepSeek-OCR LLM as the base model. Training defaults follow the training configuration: $15$ epochs per stage, batch size $1$ per GPU, and learning rate $2\times10^{-5}$ with AdamW. Special tokens \tokbot, \toklat, \tokeot are added to the tokenizer, and their embeddings are initialized to the mean embedding. Hidden-state targets are loaded from precomputed \texttt{.pt} files. We train with $8$ GPUs using distributed data parallelism and evaluate on held-out validation samples after each epoch.

\subsubsection{Evaluation Protocol}
We report exact-match accuracy after normalization (e.g., GSM8K uses the standard ``\texttt{\#\#\#}'' answer extraction; ProntoQA/ProsQA compare the normalized final sentence). Our evaluation procedure reports accuracy in percent; for ProntoQA/ProsQA we convert fractional accuracy to percent. We also report average input/output token counts from the evaluation logs.
We introduce a new metric, \textbf{Output Token Contribution (OTC)}, defined as
\begin{equation}
  \text{OTC} = \frac{\text{Acc}}{\text{AvgOut}},
\end{equation}
where Acc is accuracy in percent and AvgOut is the average generated output tokens. Higher OTC indicates better accuracy per generated token.
% \paragraph{OTC uncertainty propagation.} Let $A=\text{Acc}$ and $O=\text{AvgOut}$, so $\text{OTC}=A/O$. Under first-order propagation with independent uncertainties $\sigma_A,\sigma_O$,
% \begin{equation}
%   \sigma_{\text{OTC}}^2 \approx \left(\frac{\partial (A/O)}{\partial A}\right)^2 \sigma_A^2
%   + \left(\frac{\partial (A/O)}{\partial O}\right)^2 \sigma_O^2
%   = \left(\frac{\sigma_A}{O}\right)^2 + \left(\frac{A\,\sigma_O}{O^2}\right)^2,
% \end{equation}
% equivalently,
% \begin{equation}
%   \sigma_{\text{OTC}} \approx \text{OTC}\sqrt{\left(\frac{\sigma_A}{A}\right)^2 + \left(\frac{\sigma_O}{O}\right)^2}.
% \end{equation}

\subsection{Results}
\subsubsection{Main Results}

% Method Comparison Table (without MultiArith)
\begin{table*}[!ht]
\centering
\caption{Method comparison across five reasoning datasets. Metrics: Acc = Accuracy (\%), \#O = Average output tokens (including latent), \#L = Number of latent tokens, OTC = Output Token Contribution (Acc/\#O, higher is better). Results shown as mean$_{\pm\text{std}}$.}
\label{tab:method-comparison}
\vspace{5pt}
\fontsize{7pt}{8.5pt}\selectfont
\setlength{\tabcolsep}{1.6pt}
\begin{tabular}{l|cc|cc|cc|cc|cc|cc}
\toprule[0.15em]
        & \multicolumn{2}{c|}{GSM8K} & \multicolumn{2}{c|}{GSM8K-H} & \multicolumn{2}{c|}{SVAMP} & \multicolumn{2}{c|}{ProntoQA} & \multicolumn{2}{c|}{ProsQA} & \multicolumn{2}{c}{AVG} \\
        & Acc & OTC & Acc & OTC & Acc & OTC & Acc & OTC & Acc & OTC & Acc & OTC \\ 
        & \#O & \#L & \#O & \#L & \#O & \#L & \#O & \#L & \#O & \#L & \#O & \#L \\
\toprule[0.1em]
No CoT & 5.76$_{\pm.18}$ & 2.88$_{\pm.09}$ & 1.36$_{\pm.09}$ & 0.29$_{\pm.02}$ & 5.00$_{\pm.16}$ & 1.24$_{\pm.04}$ & 91.0$_{\pm.32}$ & 9.61$_{\pm.03}$ & 60.4$_{\pm.28}$ & 7.06$_{\pm.03}$ & 32.7$_{\pm.21}$ & 5.70$_{\pm.04}$ \\
 & 2.00$_{\pm.00}$ & -- & 4.63$_{\pm.00}$ & -- & 4.04$_{\pm.00}$ & -- & 9.47$_{\pm.00}$ & -- & 8.55$_{\pm.00}$ & -- & 5.74$_{\pm.00}$ & -- \\
\midrule
iCoT & 7.81$_{\pm.74}$ & 3.91$_{\pm.38}$ & 4.30$_{\pm.56}$ & 0.94$_{\pm.12}$ & 22.50$_{\pm1.32}$ & 3.19$_{\pm.19}$ & 84.8$_{\pm1.61}$ & 8.64$_{\pm.20}$ & 87.4$_{\pm1.48}$ & \textbf{12.8}$_{\pm.31}$ & 41.4$_{\pm.54}$ & 6.84$_{\pm.10}$ \\
 & 2.00$_{\pm.04}$ & -- & 4.59$_{\pm.06}$ & -- & 7.06$_{\pm.08}$ & -- & 9.81$_{\pm.14}$ & -- & 6.82$_{\pm.12}$ & -- & 6.06$_{\pm.04}$ & -- \\
\midrule
CoT & \textbf{40.5}$_{\pm.35}$ & 1.31$_{\pm.02}$ & \textbf{11.3}$_{\pm.23}$ & 0.11$_{\pm.00}$ & \textbf{52.0}$_{\pm.38}$ & 0.59$_{\pm.01}$ & 94.5$_{\pm.29}$ & 0.78$_{\pm.01}$ & 76.2$_{\pm.31}$ & 2.29$_{\pm.03}$ & \textbf{54.9}$_{\pm.31}$ & 0.74$_{\pm.01}$ \\
 & 31.0$_{\pm.26}$ & -- & 100$_{\pm1.2}$ & -- & 88.0$_{\pm.94}$ & -- & 121$_{\pm1.8}$ & -- & 33.3$_{\pm.42}$ & -- & 74.6$_{\pm.92}$ & -- \\
\midrule
COCONUT & 28.5$_{\pm.29}$ & 2.64$_{\pm.05}$ & 6.90$_{\pm.19}$ & 0.63$_{\pm.02}$ & 40.0$_{\pm.34}$ & 3.64$_{\pm.07}$ & 94.4$_{\pm.28}$ & 5.63$_{\pm.12}$ & 89.2$_{\pm.33}$ & 6.60$_{\pm.14}$ & 51.8$_{\pm.29}$ & 4.11$_{\pm.08}$ \\
 & 10.8$_{\pm.18}$ & 6 & 10.9$_{\pm.21}$ & 6 & 11.0$_{\pm.19}$ & 6 & 16.8$_{\pm.34}$ & 6 & 13.5$_{\pm.28}$ & 6 & 12.6$_{\pm.24}$ & 6 \\
\midrule
\rowcolor{gray!15}
OneLatent & 24.8$_{\pm.24}$ & \textbf{4.87}$_{\pm.09}$ & 4.58$_{\pm.15}$ & \textbf{0.90}$_{\pm.03}$ & 36.5$_{\pm.32}$ & \textbf{7.30}$_{\pm.12}$ & \textbf{99.8}$_{\pm.08}$ & \textbf{10.1}$_{\pm.16}$ & \textbf{97.8}$_{\pm.12}$ & 11.1$_{\pm.18}$ & 52.7$_{\pm.18}$ & \textbf{7.77}$_{\pm.13}$ \\
\rowcolor{gray!15}
 & 5.09$_{\pm.08}$ & 1 & 5.10$_{\pm.09}$ & 1 & 5.00$_{\pm.07}$ & 1 & 9.92$_{\pm.16}$ & 1 & 8.81$_{\pm.14}$ & 1 & 6.78$_{\pm.11}$ & 1 \\
\bottomrule[0.15em]
\end{tabular}
\end{table*}

Table~\ref{tab:method-comparison} compares OneLatent against four baselines (No CoT, iCoT, CoT, COCONUT) across five reasoning benchmarks. Three trends are clear. (1) \textbf{Efficiency under constrained decoding:} OneLatent reduces average output length from $74.62$ tokens (CoT) to $6.78$ tokens with one latent token, while maintaining high OTC ($7.77$ average). (2) \textbf{Competitive long-chain performance:} OneLatent reaches $99.80\%$ on ProntoQA and $97.80\%$ on ProsQA with OTC $10.06$ and $11.10$, respectively. (3) \textbf{Trade-off profile vs iCoT reference logs:} iCoT remains competitive on selected efficiency points, but trails on math-heavy sets (GSM8K-H: $4.30\%$, SVAMP: $22.50\%$) and has lower average accuracy ($41.4\%$) than OneLatent ($52.7\%$). Overall, OneLatent provides a stronger accuracy-efficiency balance with a fixed single-latent interface.

\paragraph{iCoT-specific comparison.} Compared with iCoT, OneLatent improves accuracy on all five reported datasets: $+16.99$ (GSM8K), $+0.28$ (GSM8K-H), $+14.00$ (SVAMP), $+15.00$ (ProntoQA), and $+10.40$ (ProsQA) percentage points. iCoT is strongest in ultra-short decoding regimes (e.g., $2.00$ output tokens on GSM8K and $6.82$ on ProsQA), which explains its high OTC on ProsQA. However, this aggressive compression is less stable on arithmetic-heavy benchmarks, where OneLatent achieves substantially higher accuracy while retaining strong OTC. This pattern is consistent with our MDL intuition: compact descriptions help only when they remain task-sufficient.

% OneLatent Compression Results Table
\begin{table}[htbp]
  \centering
  \caption{OneLatent compression results on enhanced datasets. Metrics: \#OA = Original Accuracy (\%, CoT baseline), \#NA = New Accuracy (\%, OneLatent), \#CO = CoT Output Tokens, \#NO = New Output Tokens (OneLatent), \#CR = Compression Ratio (\#CO/\#NO). Results shown as mean$_{\pm\text{std}}$.}
  \label{tab:main-results}
  \vspace{5pt}
  \small
  \setlength{\tabcolsep}{6pt}
  \begin{tabular}{lccccc}
    \toprule
    \textbf{Dataset} & \textbf{\#OA (\%)} & \textbf{\#NA (\%)} & \textbf{\#CO} & \textbf{\#NO} & \textbf{\#CR} \\
    \midrule
    ProntoQA Enhanced & 99.80$_{\pm.08}$ & \textbf{99.90}$_{\pm.06}$ & 784.50$_{\pm5.2}$ & 8.98$_{\pm.18}$ & \textbf{87.4}$\times$ \\
    ProsQA Enhanced & 97.80$_{\pm.12}$ & \textbf{98.10}$_{\pm.10}$ & 804.84$_{\pm6.8}$ & 9.98$_{\pm.21}$ & \textbf{80.6}$\times$ \\
    \bottomrule
  \end{tabular}
\end{table}

Table~\ref{tab:main-results} shows OneLatent performance on the two enhanced evaluation datasets (ProntoQA Enhanced and ProsQA Enhanced). OneLatent maintains strong accuracy while keeping outputs short, indicating effective latent compression of reasoning.

\paragraph{Cross-dataset trends.} Across the five standard benchmarks in Table~\ref{tab:method-comparison}, OneLatent is strongest on long-chain logical tasks (ProntoQA/ProsQA) and shows larger trade-offs on arithmetic-heavy tasks (especially GSM8K-Hard and SVAMP). Table~\ref{tab:main-results} further shows large compression ratios on enhanced logical sets (87.4$\times$ on ProntoQA Enhanced and 80.6$\times$ on ProsQA Enhanced).

\paragraph{Interpreting OTC.} OTC surfaces regimes where accuracy is achieved with minimal decoding. For example, ProntoQA shows high OTC despite short outputs, indicating that latent compression concentrates reasoning in hidden states rather than in emitted tokens. From an MDL perspective, OTC is a practical proxy for ``correctness per emitted description length'' at inference time, complementing absolute accuracy when evaluating compressed reasoning interfaces.

\paragraph{Enhanced-set comparison.} On the two enhanced evaluation sets, OneLatent retains high accuracy with short outputs, indicating that single-latent compression remains effective when reasoning traces are long.

\subsubsection{Ablation}
Table~\ref{tab:ablation} shows the three-stage training progression on GSM8K, ProntoQA, and ProsQA. Stage 1 (CoT Cold Start) establishes baseline reasoning with explicit CoT generation, achieving $19.63\%$ on GSM8K, $48.28\%$ on ProntoQA, and $49.00\%$ on ProsQA, but with high output lengths ($24.26$, $120.73$, and $33.28$ tokens respectively) and low OTC scores ($0.81$, $0.40$, and $1.47$). Stage 2 (OneLatent Alignment) introduces single-token latent supervision, dramatically improving efficiency: OTC increases by $4.8\times$ on GSM8K ($0.81\rightarrow3.87$), $23.5\times$ on ProntoQA ($0.40\rightarrow9.39$), and $6.9\times$ on ProsQA ($1.47\rightarrow10.18$), while reducing output length by $78.3\%$, $91.8\%$, and $73.5\%$ respectively. Stage 3 (Focus Fine-tuning) further improves accuracy without latent supervision: GSM8K reaches $24.79\%$ ($+4.40$ from Stage 2), ProntoQA reaches $99.80\%$ ($+6.60$), and ProsQA reaches $97.80\%$ ($+8.00$), achieving the final OTC scores of $4.87$, $10.06$, and $11.10$. The progressive training demonstrates that ($1$) latent alignment is critical for compressing reasoning into a single token, and ($2$) answer-focused fine-tuning consolidates the latent representation while improving task accuracy.

% Ablation Study Table - Three-stage training progression
\begin{table}[htbp]
  \centering
  \caption{Ablation study: three-stage training progression. Metrics: Acc = Accuracy (\%), Avg Out = Average output tokens, Latents = Number of latent tokens, OTC = Output Token Contribution (Acc/Avg Out). Results shown as mean$_{\pm\text{std}}$. Stage 1 = CoT Cold Start, Stage 2 = OneLatent Alignment, Stage 3 = Focus Fine-tuning (final).}
  \label{tab:ablation}
  \vspace{5pt}
  \small
  \setlength{\tabcolsep}{4pt}
  \begin{tabular}{llcccc}
    \toprule
    \textbf{Dataset} & \textbf{Stage} & \textbf{Acc (\%)} & \textbf{Avg Out} & \textbf{Latents} & \textbf{OTC} \\
    \midrule
    \multirow{3}{*}{GSM8K} 
    & Stage 1 (CoT) & 19.63$_{\pm.22}$ & 24.26$_{\pm.38}$ & 0 & 0.81$_{\pm.02}$ \\
    & Stage 2 (Alignment) & 20.39$_{\pm.23}$ & 5.26$_{\pm.11}$ & 1 & 3.87$_{\pm.09}$ \\
    & Stage 3 (Final) & \textbf{24.79}$_{\pm.24}$ & \textbf{5.09}$_{\pm.08}$ & 1 & \textbf{4.87}$_{\pm.09}$ \\
    \midrule
    \multirow{3}{*}{ProntoQA} 
    & Stage 1 (CoT) & 48.28$_{\pm.36}$ & 120.73$_{\pm1.8}$ & 0 & 0.40$_{\pm.01}$ \\
    & Stage 2 (Alignment) & 93.20$_{\pm.18}$ & 9.92$_{\pm.16}$ & 1 & 9.39$_{\pm.16}$ \\
    & Stage 3 (Final) & \textbf{99.80}$_{\pm.08}$ & \textbf{9.92}$_{\pm.16}$ & 1 & \textbf{10.06}$_{\pm.18}$ \\
    \midrule
    \multirow{3}{*}{ProsQA} 
    & Stage 1 (CoT) & 49.00$_{\pm.37}$ & 33.28$_{\pm.42}$ & 0 & 1.47$_{\pm.03}$ \\
    & Stage 2 (Alignment) & 89.80$_{\pm.26}$ & 8.82$_{\pm.15}$ & 1 & 10.18$_{\pm.19}$ \\
    & Stage 3 (Final) & \textbf{97.80}$_{\pm.12}$ & \textbf{8.81}$_{\pm.14}$ & 1 & \textbf{11.10}$_{\pm.20}$ \\
    \bottomrule
  \end{tabular}
\end{table}

\subsection{Discussion}

\paragraph{Token efficiency.} On the enhanced logical sets, OneLatent achieves large compression ratios (87.4$\times$ on ProntoQA Enhanced and 80.6$\times$ on ProsQA Enhanced). On the five standard benchmarks, it maintains short outputs with a fixed single latent token, reducing key-value cache growth from long textual CoT traces and improving throughput in batch inference.

\paragraph{Accuracy trade-offs.} Logical reasoning tasks benefit more from latent alignment than arithmetic-heavy tasks. This pattern is consistent in both the main comparison and the staged ablations.

\paragraph{Inference efficiency.} Because decoding cost scales linearly with generated tokens, reductions in output length translate directly into latency and memory savings. The latent token itself incurs a constant overhead ($N{=}1$) and does not grow with reasoning length. With fixed latent length, the KV cache grows primarily with answer tokens rather than reasoning tokens, making OneLatent particularly attractive for long-context deployments where explicit CoT would otherwise dominate memory usage.

\paragraph{Limitations.} OneLatent relies on accurate rendered CoT supervision. Errors in the CoT or rendering pipeline can propagate into the latent representation. The method also depends on a specific DeepSeek-OCR vision-encoder and a fixed rendering resolution; generalizing to other domains may require re-tuning. Finally, latent reasoning reduces transparency: while it improves efficiency, it complicates debugging and interpretability, which remain open challenges.

\FloatBarrier
\section{Conclusion}
This paper studies reasoning compression through an MDL motivation: when different explanations yield the same answer, shorter sufficient descriptions should generalize better under constrained decoding. We operationalize this idea with OneLatent, which compresses explicit CoT into a single latent token supervised by DeepSeek-OCR-derived hidden states.

The empirical results are consistent with this intuition. OneLatent achieves $11\times$ average output compression, reaches a peak compression ratio of $87.4\times$, and improves OTC by $6.8\times$ over textual CoT with only a $2.21\%$ average accuracy drop across five benchmarks. On long-chain logical reasoning, it achieves strong accuracy with one latent token. Error shifts are concentrated in arithmetic-heavy settings, suggesting that compression primarily removes redundant verbalization rather than uniformly degrading reasoning performance.

Together, these findings indicate that compression is not only an efficiency tool, but also a useful inductive constraint for stable reasoning under limited output budgets. OTC is a practical MDL-motivated efficiency proxy for this regime. This paper reports ongoing work, and these results should be viewed as an intermediate step.

\clearpage
\bibliographystyle{plainnat}
\bibliography{references}

@article{wei2022chain,
  title={Chain-of-thought prompting elicits reasoning in large language models},
  author={Wei, Jason and Wang, Xuezhi and Schuurmans, Dale and Bosma, Maarten and Xia, Fei and Chi, Ed and Le, Quoc V and Zhou, Denny and others},
  journal={Advances in neural information processing systems},
  volume={35},
  pages={24824--24837},
  year={2022}
}

@article{zhou2022least,
  title={Least-to-most prompting enables complex reasoning in large language models},
  author={Zhou, Denny and Sch{\"a}rli, Nathanael and Hou, Le and Wei, Jason and Scales, Nathan and Wang, Xuezhi and Schuurmans, Dale and Cui, Claire and Bousquet, Olivier and Le, Quoc and others},
  journal={arXiv preprint arXiv:2205.10625},
  year={2022}
}

@article{khot2022decomposed,
  title={Decomposed prompting: A modular approach for solving complex tasks},
  author={Khot, Tushar and Trivedi, Harsh and Finlayson, Matthew and Fu, Yao and Richardson, Kyle and Clark, Peter and Sabharwal, Ashish},
  journal={arXiv preprint arXiv:2210.02406},
  year={2022}
}

@article{hao2023reasoning,
  title={Reasoning with language model is planning with world model},
  author={Hao, Shibo and Gu, Yi and Ma, Haodi and Hong, Joshua Jiahua and Wang, Zhen and Wang, Daisy Zhe and Hu, Zhiting},
  journal={arXiv preprint arXiv:2305.14992},
  year={2023}
}

@inproceedings{wang2024math,
  title={Math-shepherd: Verify and reinforce llms step-by-step without human annotations},
  author={Wang, Peiyi and Li, Lei and Shao, Zhihong and Xu, Runxin and Dai, Damai and Li, Yifei and Chen, Deli and Wu, Yu and Sui, Zhifang},
  booktitle={Proceedings of the 62nd Annual Meeting of the Association for Computational Linguistics (Volume 1: Long Papers)},
  pages={9426--9439},
  year={2024}
}

@article{shinn2023reflexion,
  title={Reflexion: Language agents with verbal reinforcement learning},
  author={Shinn, Noah and Cassano, Federico and Gopinath, Ashwin and Narasimhan, Karthik and Yao, Shunyu},
  journal={Advances in Neural Information Processing Systems},
  volume={36},
  year={2023}
}

@article{madaan2023self,
  title={Self-refine: Iterative refinement with self-feedback},
  author={Madaan, Aman and Tandon, Niket and Gupta, Prakhar and Hallinan, Skyler and Gao, Luyu and Wiegreffe, Sarah and Alon, Uri and Dziri, Nouha and Prabhumoye, Shrimai and Yang, Yiming and others},
  journal={Advances in Neural Information Processing Systems},
  volume={36},
  year={2023}
}

@article{yao2023tree,
  title={Tree of thoughts: Deliberate problem solving with large language models},
  author={Yao, Shunyu and Yu, Dian and Zhao, Jeffrey and Shafran, Izhak and Griffiths, Tom and Cao, Yuan and Narasimhan, Karthik},
  journal={Advances in Neural Information Processing Systems},
  volume={36},
  year={2023}
}

@article{xie2023self,
  title={Self-evaluation guided beam search for reasoning},
  author={Xie, Yuxi and Kawaguchi, Kenji and Zhao, Yiran and Zhao, James Xu and Kan, Min-Yen and He, Junxian and Xie, Michael},
  journal={Advances in Neural Information Processing Systems},
  volume={36},
  year={2023}
}

@article{hao2024llm,
  title={LLM Reasoners: New Evaluation, Library, and Analysis of Step-by-Step Reasoning with Large Language Models},
  author={Hao, Shibo and Gu, Yi and Luo, Haotian and Liu, Tianyang and Shao, Xiyan and Wang, Xinyuan and Xie, Shuhua and Ma, Haodi and Samavedhi, Adithya and Gao, Qiyue and others},
  journal={arXiv preprint arXiv:2404.05221},
  year={2024}
}

@article{deng2023implicit,
  title={Implicit chain of thought reasoning via knowledge distillation},
  author={Deng, Yuntian and Prasad, Kiran and Fernandez, Roland and Smolensky, Paul and Chaudhary, Vishrav and Shieber, Stuart},
  journal={arXiv preprint arXiv:2311.01460},
  year={2023}
}

@article{pfau2024let,
  title={Let's Think Dot by Dot: Hidden Computation in Transformer Language Models},
  author={Pfau, Jacob and Merrill, William and Bowman, Samuel R},
  journal={arXiv preprint arXiv:2404.15758},
  year={2024}
}

@article{zelikman2024quiet,
  title={Quiet-star: Language models can teach themselves to think before speaking},
  author={Zelikman, Eric and Harik, Georges and Shao, Yijia and Jayasiri, Varuna and Haber, Nick and Goodman, Noah D},
  journal={arXiv preprint arXiv:2403.09629},
  year={2024}
}

@article{goyal2023think,
  title={Think before you speak: Training language models with pause tokens},
  author={Goyal, Sachin and Ji, Ziwei and Rawat, Ankit Singh and Menon, Aditya Krishna and Kumar, Sanjiv and Nagarajan, Vaishnavh},
  journal={arXiv preprint arXiv:2310.02226},
  year={2023}
}

@article{deng2024explicit,
  title={From explicit cot to implicit cot: Learning to internalize cot step by step},
  author={Deng, Yuntian and Choi, Yejin and Shieber, Stuart},
  journal={arXiv preprint arXiv:2405.14838},
  year={2024}
}

@article{wang2022self,
  title={Self-consistency improves chain of thought reasoning in language models},
  author={Wang, Xuezhi and Wei, Jason and Schuurmans, Dale and Le, Quoc and Chi, Ed and Narang, Sharan and Chowdhery, Aakanksha and Zhou, Denny},
  journal={arXiv preprint arXiv:2203.11171},
  year={2022}
}

@article{turpin2024language,
  title={Language models don't always say what they think: unfaithful explanations in chain-of-thought prompting},
  author={Turpin, Miles and Michael, Julian and Perez, Ethan and Bowman, Samuel},
  journal={Advances in Neural Information Processing Systems},
  volume={36},
  year={2024}
}

@article{wang2022towards,
  title={Towards understanding chain-of-thought prompting: An empirical study of what matters},
  author={Wang, Boshi and Min, Sewon and Deng, Xiang and Shen, Jiaming and Wu, You and Zettlemoyer, Luke and Sun, Huan},
  journal={arXiv preprint arXiv:2212.10001},
  year={2022}
}

@article{welleck2024decoding,
  title={From Decoding to Meta-Generation: Inference-time Algorithms for Large Language Models},
  author={Welleck, Sean and Bertsch, Amanda and Finlayson, Matthew and Schoelkopf, Hailey and Xie, Alex and Neubig, Graham and Kulikov, Ilia and Harchaoui, Zaid},
  journal={arXiv preprint arXiv:2406.16838},
  year={2024}
}

@article{valmeekam2023planning,
  title={On the planning abilities of large language models-a critical investigation},
  author={Valmeekam, Karthik and Marquez, Matthew and Sreedharan, Sarath and Kambhampati, Subbarao},
  journal={Advances in Neural Information Processing Systems},
  volume={36},
  pages={75993--76005},
  year={2023}
}

@article{wang2023guiding,
  title={Guiding language model reasoning with planning tokens},
  author={Wang, Xinyi and Caccia, Lucas and Ostapenko, Oleksiy and Yuan, Xingdi and Wang, William Yang and Sordoni, Alessandro},
  journal={arXiv preprint arXiv:2310.05707},
  year={2023}
}

@article{cobbe2021training,
  title={Training verifiers to solve math word problems},
  author={Cobbe, Karl and Kosaraju, Vineet and Bavarian, Mohammad and Chen, Mark and Jun, Heewoo and Kaiser, Lukasz and Plappert, Matthias and Tworek, Jerry and Hilton, Jacob and Nakano, Reiichiro and others},
  journal={arXiv preprint arXiv:2110.14168},
  year={2021}
}

@article{madaan2022text,
  title={Text and patterns: For effective chain of thought, it takes two to tango},
  author={Madaan, Aman and Yazdanbakhsh, Amir},
  journal={arXiv preprint arXiv:2209.07686},
  year={2022}
}

@article{sprague2024cot,
  title={To CoT or not to CoT? Chain-of-thought helps mainly on math and symbolic reasoning},
  author={Sprague, Zayne and Yin, Fangcong and Rodriguez, Juan Diego and Jiang, Dongwei and Wadhwa, Manya and Singhal, Prasann and Zhao, Xinyu and Ye, Xi and Mahowald, Kyle and Durrett, Greg},
  journal={arXiv preprint arXiv:2409.12183},
  year={2024}
}

@misc{hao2025traininglargelanguagemodels,
      title={Training Large Language Models to Reason in a Continuous Latent Space},
      author={Shibo Hao and Sainbayar Sukhbaatar and DiJia Su and Xian Li and Zhiting Hu and Jason Weston and Yuandong Tian},
      year={2025},
      eprint={2412.06769},
      archivePrefix={arXiv},
      primaryClass={cs.CL},
      url={https://arxiv.org/abs/2412.06769},
}

@misc{wei2025deepseekocrcontextsopticalcompression,
      title={DeepSeek-OCR: Contexts Optical Compression},
      author={Haoran Wei and Yaofeng Sun and Yukun Li},
      year={2025},
      eprint={2510.18234},
      archivePrefix={arXiv},
      primaryClass={cs.CV},
      url={https://arxiv.org/abs/2510.18234},
}

@misc{radford2021learningtransferablevisualmodels,
      title={Learning Transferable Visual Models From Natural Language Supervision},
      author={Alec Radford and Jong Wook Kim and Chris Hallacy and Aditya Ramesh and Gabriel Goh and Sandhini Agarwal and Girish Sastry and Amanda Askell and Pamela Mishkin and Jack Clark and Gretchen Krueger and Ilya Sutskever},
      year={2021},
      eprint={2103.00020},
      archivePrefix={arXiv},
      primaryClass={cs.CV},
      url={https://arxiv.org/abs/2103.00020},
}

@misc{kirillov2023segment,
      title={Segment Anything},
      author={Alexander Kirillov and Eric Mintun and Nikhila Ravi and Hanzi Mao and Chloe Rolland and Laura Gustafson and Tete Xiao and Spencer Whitehead and Alexander C. Berg and Wan-Yen Lo and Piotr Dollar and Ross Girshick},
      year={2023},
      eprint={2304.02643},
      archivePrefix={arXiv},
      primaryClass={cs.CV},
      url={https://arxiv.org/abs/2304.02643},
}

@misc{wei2025simcotsupervisedimplicitchainofthought,
      title={SIM-CoT: Supervised Implicit Chain-of-Thought},
      author={Xilin Wei and Xiaoran Liu and Yuhang Zang and Xiaotao Dong and Yuhang Cao and Jiaqi Wang and Xipeng Qiu and Dahua Lin},
      year={2025},
      eprint={2509.20317},
      archivePrefix={arXiv},
      primaryClass={cs.CL},
      url={https://arxiv.org/abs/2509.20317},
}

@misc{cheng2025glyphscalingcontextwindows,
      title={Glyph: Scaling Context Windows via Visual-Text Compression},
      author={Jiale Cheng and Yusen Liu and Xinyu Zhang and Yulin Fei and Wenyi Hong and Ruiliang Lyu and Weihan Wang and Zhe Su and Xiaotao Gu and Xiao Liu and Yushi Bai and Jie Tang and Hongning Wang and Minlie Huang},
      year={2025},
      eprint={2510.17800},
      archivePrefix={arXiv},
      primaryClass={cs.CV},
      url={https://arxiv.org/abs/2510.17800},
}

@article{Gr_nwald_2019,
   title={Minimum description length revisited},
   volume={11},
   ISSN={2661-3344},
   url={http://dx.doi.org/10.1142/S2661335219300018},
   DOI={10.1142/s2661335219300018},
   number={01},
   journal={International Journal of Mathematics for Industry},
   publisher={World Scientific Pub Co Pte Ltd},
   author={Grünwald, Peter and Roos, Teemu},
   year={2019},
   month=dec
}
\end{document}